%% file: root.tex
\documentclass[lettersize,journal]{IEEEtran}
\usepackage{booktabs}
\usepackage{amsmath,amsfonts}
\usepackage{algorithmic}
\usepackage{algorithm}
\usepackage{array}
\usepackage{subcaption}
\usepackage{textcomp}
\usepackage{stfloats}
\usepackage{pifont}    % for \ding{55}
\usepackage{url}
\usepackage{multirow}
\usepackage{lipsum}
\usepackage{graphicx}
\usepackage{makecell} 
\usepackage{verbatim}
\usepackage{graphicx}
\usepackage{cite}
\usepackage{graphicx}
\usepackage{subcaption}
\usepackage{caption}
\usepackage{amsmath}
\usepackage{array}
\usepackage{colortbl}
\usepackage[table,xcdraw]{xcolor}
\usepackage{arydshln}
\usepackage{caption}

\newcommand{\redx}{{\color{red}\ding{55}}}

\definecolor{myteal}{HTML}{95CEC7}
\definecolor{myyellow}{HTML}{FAF3B3}
\definecolor{mycoral}{HTML}{F8CA8C}

\begin{document}
\pagestyle{empty}

 \title{GRAND-SLAM: Local Optimization for Globally Consistent \\ Large-Scale Multi-Agent Gaussian SLAM}

 % \title{GRAND-SLAM: \textbf{Co}llabo\textbf{r}ative \textbf{G}auss\textbf{i}an SLAM for  \\ Large-Scale Outdoor Environments}

\author{Annika Thomas, Aneesa Sonawalla, Alex Rose, Jonathan P. How
        % <-this % stops a space
%\thanks{This paper was produced by the IEEE Publication Technology Group. They are in Piscataway, NJ.}% <-this % stops a space
\thanks{Authors are with the Aerospace Controls Laboratory at Massachusetts Institute of Technology Department of Aeronautics and Astronautics}
}

\maketitle
\thispagestyle{empty}

\input{paper/abstract}

% \begin{IEEEkeywords}
% Localization, Mapping, Gaussian Splatting, RGB-D SLAM, Multi-Agent SLAM
% \end{IEEEkeywords}

\input{paper/introduction}
\input{paper/related_works}
\input{paper/method}
\input{paper/results}

\input{paper/conclusion}

% \section*{Acknowledgments}
% This work is supported in part by the Ford Motor Company, ONR, and ARL DCIST under Cooperative Agreement Number W911NF-17-2-0181.


% \balance
% \bibliographystyle{ieeetr}
\bibliographystyle{IEEEtran} % was IEEEtran, added 'N' to support natbib
\bibliography{references}

\end{document}

%% file: paper/abstract.tex
\begin{abstract}
    3D Gaussian splatting has emerged as an expressive scene representation for RGB-D visual SLAM, but its application to large-scale, multi-agent outdoor environments remains unexplored. Multi-agent Gaussian SLAM is a promising approach to rapid exploration and reconstruction of environments, offering scalable environment representations, but existing approaches are limited to small-scale, indoor environments. To that end, we propose Gaussian Reconstruction via Multi-Agent Dense SLAM, or GRAND-SLAM, a collaborative Gaussian splatting SLAM method that integrates i) an implicit tracking module based on local optimization over submaps and ii) an approach to inter- and intra-robot loop closure integrated into a pose-graph optimization framework. Experiments show that GRAND-SLAM provides state-of-the-art tracking performance and 28\% higher PSNR than existing methods on the Replica indoor dataset, as well as 91\% lower multi-agent tracking error and improved rendering over existing multi-agent methods on the large-scale, outdoor Kimera-Multi dataset. 
\end{abstract}

% Demonstrations on the Replica Multiagent simulated indoor dataset show that GRAND-SLAM outperforms existing methods in single-agent tracking with and without loop closure and provides better rendering performance than existing methods. Experiments on the Kimera Multi dataset demonstrate that GRAND-SLAM outperforms existing methods for multi-agent operations in large-scale, outdoor environments in both tracking and rendering performance.  

%% file: paper/introduction.tex
\section{Introduction}

\label{sec:intro}

% Motivation! Start with definition of SLAM and usefulness.
% Background and Evolution of Visual SLAM in Diverse Applications
Visual Simultaneous Localization and Mapping (SLAM) is a foundational technology for a variety of applications including real-time spatial awareness for Virtual Reality and Augmented Reality (AR/VR), autonomous driving, and robot navigation. In these contexts, visual SLAM allows systems to localize themselves within an environment while creating maps that can support complex tasks like navigation, object interaction and scene understanding. SLAM has been an active area of development for the past two decades \cite{cadena2016past, fuentes2015visual, kazerouni2022survey}, and many approaches have been developed using sparse scene representations with point clouds \cite{davison2007monoslam, klein2007parallel, mur2017orb}, surfels \cite{schops2019bad, whelan2015elasticfusion} and voxels \cite{newcombe2011kinectfusion}. While these methods achieve accurate tracking and real-time mapping, they struggle to render high-quality textures and are limited in capabilities for novel view synthesis.

% Emergence of 3D Gaussian Splatting (3DGS) as a Promising Representation
More recently, the development of neural implicit methods \cite{sucar2021imap, zhu2022nice, johari2023eslam, wang2023co}, particularly those leveraging neural radiance fields (NeRFs) \cite{mildenhall2021nerf}, began to address shortcomings of sparse methods by providing photorealistic map quality. 3D Gaussian splatting (3DGS) \cite{kerbl20233d} recently emerged as an efficient alternative, with comparable rendering quality to NeRFs but significantly faster rendering and training times, making it more suitable to robotics applications. Unlike NeRFs, 3D Gaussian maps can be explicitly edited or deformed, which is beneficial for tasks like map correction and adaptation in dynamic scenes. 

3DGS has recently been integrated into visual SLAM using monocular~\cite{matsuki2024gaussian} and RGB-D \cite{yugay2023gaussian, keetha2024splatam} sensing modalities, but these approaches have not been scaled to outdoor, large-scale environments. Existing visual 3DGS SLAM approaches accumulate error and map drift due to a lack of robust loop closures and global consistency mechanisms. This creates pose errors and map distortions over time, particularly when scaling the maps to large-scale or complex environments. While loop closure integration implicitly addresses the limitations of drift accumulation in large-scale environments, existing implementations \cite{liso2024loopy, zhu2024loopsplat, xu2024glc} are limited to small-scale, indoor scenes. Extension to large-scale environments in the existing literature assumes the presence of LiDAR scans to seed the mapping process \cite{wu2024hgs, hong2024liv}, but sparse scans can degrade performance and expensive, bulky LiDAR scanners may not be available. 
%In this letter, we present the first 3DGS SLAM using RGB-D for large-scale, outdoor reconstruction.

\begin{figure}[t!]
    \centering
    \includegraphics[width=\linewidth, trim={6cm, 2.5cm, 10cm, 0.02cm}, clip]{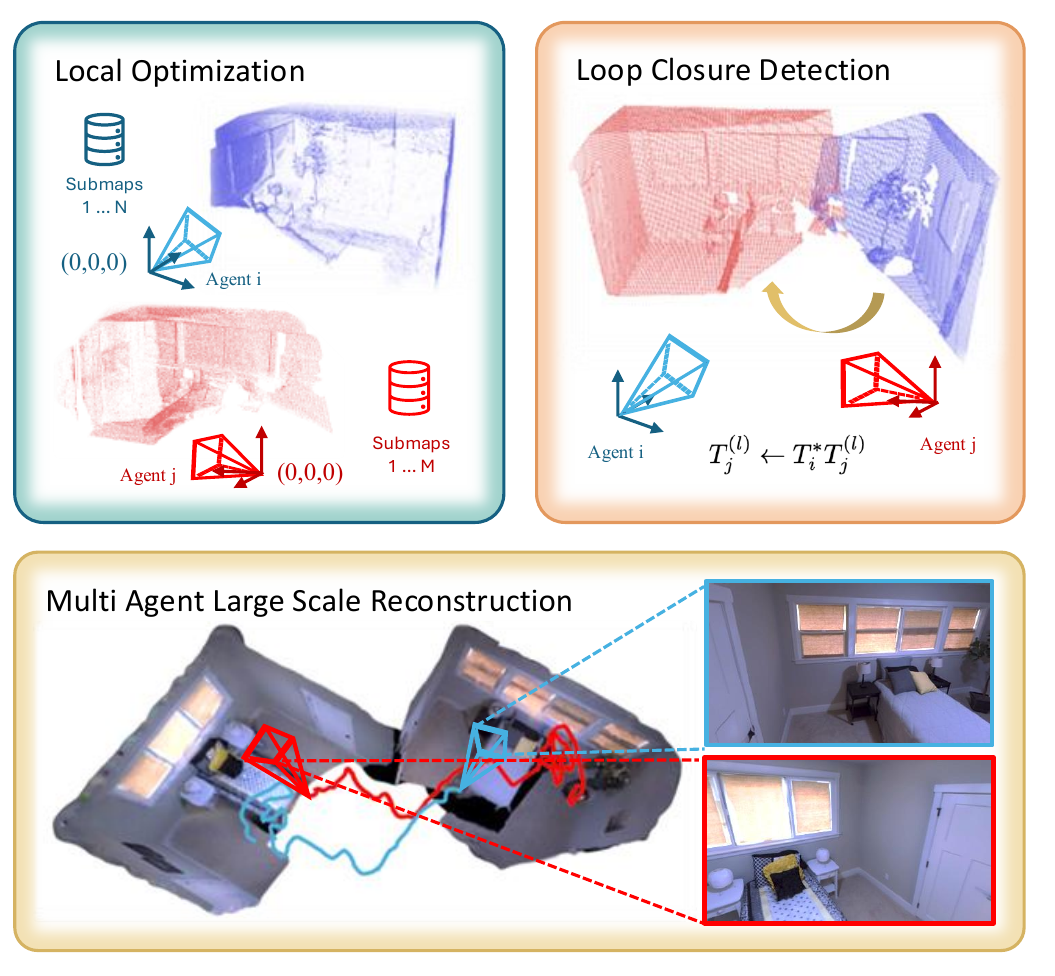}
    \caption{GRAND-SLAM introduces local optimization by submap for each agent, integrates inter-agent loop closure by submap and reconstructs large-scale environments via pose graph optimization.}
    \label{fig:first_page}
    \vspace*{-0.12in}
\end{figure}

Additionally, multi-agent SLAM is an extension of the visual SLAM problem area that holds promise for rapid exploration of novel environments while providing global consistency mechanisms. Multi-agent SLAM with traditional 3D reconstruction methods often scales poorly in large environments due to slow optimization. We address this by formulating the 3DGS SLAM problem in a way that generates compact environment representations which can be efficiently optimized in local frames, enabling deployment at scale. We combine locally optimized submaps with an explicit ICP-based registration process for inter- and intra-robot loop closures and integrate this in a distributed pose graph optimization framework, as shown in Fig. \ref{fig:first_page}.

% Proposed solution
To support robust mapping in large-scale environments, we propose GRAND-SLAM, the first method to scalably integrate inter- and intra-robot loop closures in multi-agent RGB-D Gaussian Splatting SLAM. Firstly, GRAND-SLAM incrementally builds submaps of 3D Gaussians using an RGB-D input stream for seeding and optimization. Secondly, we implement inter- and intra-robot loop closure to improve agent trajectories with a coarse-to-fine optimization process. After determining potential map overlap regions using keyframe-based descriptors, we register loop closures using a dense ICP-based approach. Finally, once loop closures are detected, the system adds nodes and edges to a pose graph and performs optimization to directly update the map based on the detected loop closures. 
Our approach demonstrates comprehensive improvements over existing RGB-D SLAM methods, delivering higher-fidelity maps, more accurate geometry, and more robust tracking in both indoor simulations and outdoor real-world settings.
% Experimental results in simulated indoor and real-world outdoor datasets demonstrate scene geometry and detail, achieving high-fidelity mapping and state-of-the-art tracking compared to other RGB-D SLAM methods.
%
The contributions of our work are: 

\begin{enumerate}
    \item A tracking module with local submap-based optimization for RGB-D Gaussian Splatting SLAM, scaling more robustly to large-scale scene representations than other radiance-field-based SLAM methods;
    \item Inter- and intra-robot loop closure detection using a coarse-to-fine optimization procedure, including an optimization procedure to jointly minimize photometric and geometric error followed by a fine ICP-based refinement;
    \item Integration of loop closures using single- and multi-agent pose graph optimization to reduce map drift over time and reduce tracking errors, as demonstrated through real-world results.
\end{enumerate}

%% file: paper/related_works.tex
\section{Related Works}
\label{sec:relatedworks}

%-------------------------------------------------------------------------
\subsection{Vision-Based Tracking and Reconstruction}
Classical visual SLAM systems, such as PTAM \cite{klein2007parallel} and MonoSLAM \cite{davison2007monoslam}, extract sparse visual features from images and track them across frames to estimate camera motion. ORB-SLAM3 \cite{campos2021orb} is a widely used feature-based SLAM approach, associating keypoints with binary descriptors to enable fast and robust data association. However, the descriptor extraction and matching process can be computationally intensive. FastORB-SLAM \cite{fu2021fast} addresses this with a lightweight approach that omits descriptor computation for more efficient tracking.

In texture-sparse scenes, point features may be unreliable. To mitigate this, alternative geometric primitives such as lines \cite{gomez2019pl} and planes \cite{li2021rgb, gong2021planefusion} have been used. Surfel-based systems like ElasticFusion \cite{whelan2015elasticfusion} and BAD-SLAM \cite{schops2019bad} enable dense map fusion and loop closure, offering continuous surface reconstruction. Voxel-based and volumetric approaches (e.g., Voxel Hashing \cite{niessner2013real}) and implicit TSDF fusion methods \cite{curless1996volumetric, newcombe2011kinectfusion, dai2017bundlefusion} have also laid important groundwork for dense tracking and mapping. Direct methods such as DTAM \cite{newcombe2011dtam} and real-time variational approaches \cite{stuhmer2010real} demonstrate the benefit of leveraging depth maps for robust camera tracking and geometry reconstruction.

% Many classic visual SLAM algorithms \cite{klein2007parallel, davison2007monoslam} extract sparse point features from images and align point clouds to track motion between different keyframes. Feature-based SLAM methods such as ORB-SLAM \cite{campos2021orb} also rely heavily on extracting sparse point features from images, but associate point features with verbal ``descriptors''. After extraction, sparse point features and associated descriptors are fed to feature matching algorithms to estimate the relative motions between viewpoints.

% Calculating descriptors can be computationally expensive. To reduce the computation in descriptor calculation, FastORB-SLAM \cite{fu2021fast} proposes a lightweight and effective method to track key points without computing descriptors. 

% Man-made scenes can be low-textured, making it difficult to provide enough reliable points to achieve robust tracking and mapping. Therefore, other types of features, such as lines \cite{gomez2019pl} and planes \cite{li2021rgb, gong2021planefusion}, are explored to compensate for the degeneration cases.

% surfels \cite{whelan2015elasticfusion, schops2019bad}

% Some stuff about voxel \cite{niessner2013real} and cuboid based SLAM

% Implicit representations: \cite{newcombe2011kinectfusion, dai2017bundlefusion}

% Some stuff about how depth cameras help with the SLAM problem. Depth map methods: \cite{newcombe2011dtam, stuhmer2010real}

%-------------------------------------------------------------------------
\subsection{Dense Neural SLAM and Implicit Representations }

Neural SLAM approaches extend dense mapping with photorealistic reconstructions by replacing explicit maps with implicit functions. NeRF \cite{mildenhall2021nerf} represents scenes as radiance fields parameterized by MLPs, enabling high-quality novel view synthesis. Systems like NICE-SLAM \cite{zhu2022nice} and Vox-Fusion \cite{yang2022vox} adapt these ideas by learning per-voxel features, while Point-SLAM \cite{sandstrom2023point} attaches learned features directly to sparse geometry. Hybrid methods such as Co-SLAM \cite{wang2023co} combine coordinate-based encodings with hash grids for efficiency.

While powerful in appearance modeling, neural SLAM pipelines often suffer from long optimization times, difficulty with real-time performance, and lack of support for rigid-body map transformations. Moreover, separating tracking from mapping can reduce drift correction capability. Loop closure techniques such as those used in Loopy-SLAM \cite{liso2024loopy} and LoopSplat \cite{zhu2024loopsplat} apply pose-graph optimization to correct trajectory errors, but performance bottlenecks remain due to the heavy reliance on learned representations.

% The seminal work of Curless and Levoy \cite{curless1996volumetric} set the stage for a variety of 3D reconstruction methods using truncated signed distance functions (TSDF).

% %-------------------------------------------------------------------------
% \subsection{NeRFs}

% (NeRF) \cite{mildenhall2021nerf} that uses implicit scene representation based on MLP networks

%-------------------------------------------------------------------------
\subsection{Gaussian Splatting SLAM}

3D Gaussian Splatting (3DGS) \cite{kerbl20233d} offers an efficient alternative to neural representations, enabling photorealistic novel view synthesis through explicit anisotropic Gaussians and differentiable rendering. Unlike NeRFs, these representations can be rendered and optimized in real time. They also natively support rigid transformations, making them particularly attractive for scene editing applications.

Recent works such as MonoGS \cite{matsuki2024gaussian} integrate 3DGS into incremental monocular SLAM. SplaTAM \cite{keetha2024splatam} further improves speed and scalability by leveraging silhouette masks for adaptive Gaussian selection. RGBD GS-ICP \cite{ha2024rgbd} replaces image-space alignment with 3D Gaussian matching, using Generalized ICP to align frames and maps while preserving rendering quality.

\subsection{Gaussian Splatting SLAM Methods with Loop Closure}
To address long-term drift in 3DGS-based SLAM, recent methods have introduced loop closure mechanisms. GLC-SLAM \cite{xu2024glc} integrates a pretrained image-based retrieval network to associate keyframe descriptors, enabling loop detection via cosine similarity. Robust GSSLAM \cite{zhu2024robust} maintains dual sets of Gaussians and computes similarity for loop detection, although practical reproducibility remains limited. MAGiC-SLAM \cite{yugay2025magic} extends 3DGS SLAM to support loop closure by leveraging a pretrained vision-language model for loop candidate selection and incorporating a differentiable alignment process, enabling drift correction and robustness in novel environments.

% Compared to prior systems, it offers improved tracking efficiency, lower runtime, and more scalable scene reconstruction.

\subsection{Multi-Agent SLAM}

Collaborative SLAM has evolved to accommodate multi-agent settings via centralized or distributed architectures. Centralized systems like CCM-SLAM \cite{schmuck2019ccm} and CVI-SLAM \cite{karrer2018cvi} use a central server for map fusion and global optimization, with individual agents performing lightweight tracking. In contrast, distributed systems such as Swarm-SLAM \cite{lajoie2023swarm} support peer-to-peer communication with robust inter-agent loop closure and increased resilience to network limitations.

More recently, neural and hybrid representations have been applied to multi-agent systems. CP-SLAM \cite{hu2023cp} incorporates NeRF-like volumetric fusion for collaborative scene reconstruction, though it remains limited in speed and agent scalability. MAGiC-SLAM \cite{yugay2025magic} builds on this direction by offering real-time multi-agent reconstruction with explicit Gaussian representations. Its centralized architecture supports multiple agents contributing partial maps and integrating them globally. Unlike NeRF-based methods, it achieves fast rendering, rigid-body map merging, and supports frequent loop closures to maintain global consistency across agents, but it is limited to small-scale indoor settings.

While CP-SLAM and MAGiC-SLAM introduce global consistency mechanisms for photorealistic SLAM in multi-agent operations, they are limited to deployment in small-scale, indoor settings. GRAND-SLAM is the first large-scale RGB-D Gaussian splatting SLAM approach that leverages inter- and intra-robot loop closure for drift reduction. It delivers higher fidelity maps and more robust tracking across small scale and large scale environments than prior work.

%% file: paper/method.tex
\section{Method}

\begin{figure*}[t!]
    \centering
    \includegraphics[width=0.98\linewidth, trim={0.5cm, 6cm, 0.5cm, 2cm}, clip]{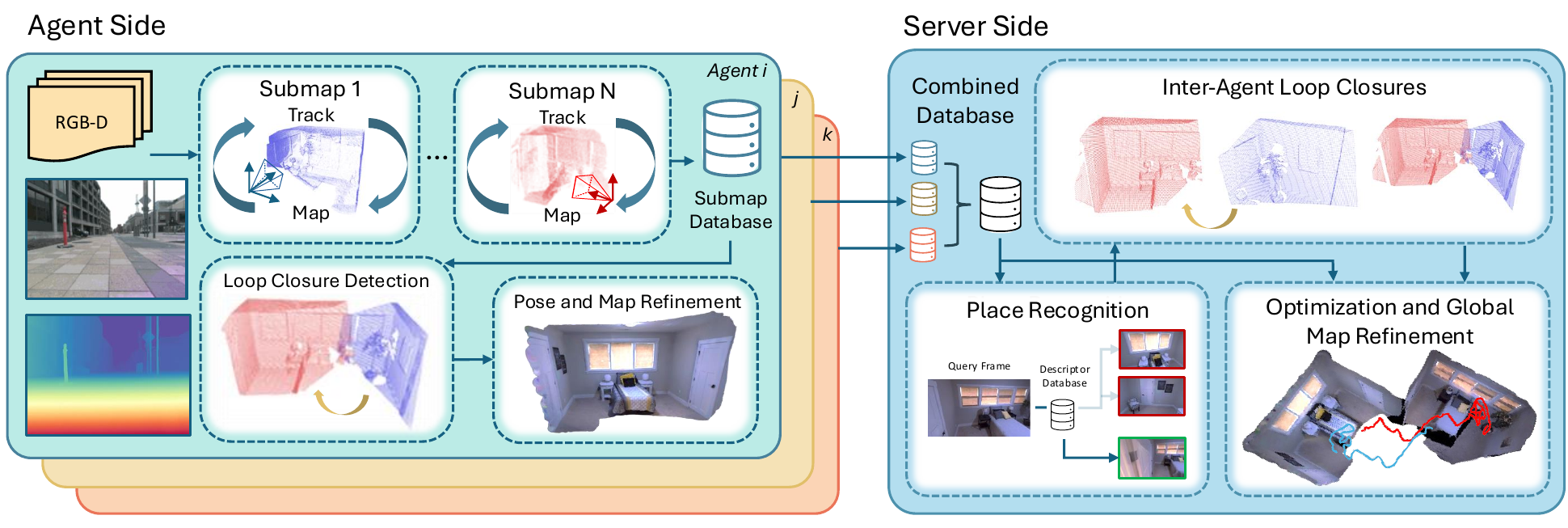}
    \caption{GRAND-SLAM can operate independently as a single-agent system or as a multi-agent system. On the agent side, RGB-D images are fed into the tracking and mapping modules for each submap, where keyframe descriptors are saved in a database for intra-agent loop closure detection. Submaps with keyframe descriptors are shared with the server side where inter-agent loop closures are detected, and the global map is refined. }
    \label{fig:enter-label}
    \vspace*{-0.1in}
\end{figure*}

\subsection{Preliminary: Gaussian Splatting}

We use 3D Gaussian splatting (3DGS) \cite{kerbl20233d} as the underlying scene representation for mapping. In contrast to implicit neural field representations, 3DGS leverages an explicit, differentiable point-based formulation, enabling real-time rendering and optimization while supporting rigid-body transformations.

Firstly, 3DGS generates a dense point cloud from an RGB-D image using camera intrinsics and per-pixel depth. To initialize the map, a set of anisotropic 3D Gaussians is seeded from the point cloud. Each Gaussian is parameterized by a position $\mu \in \mathbb{R}^3$, a 3D covariance matrix $\Sigma \in \mathbb{S}_+^3$, an opacity value $o \in \mathbb{R}$, and an RGB color vector $c \in \mathbb{R}^3$. These Gaussians are then projected onto the image plane via a differentiable rasterization pipeline, producing color and depth renderings $\hat{I}, \hat{D}$ that are compared to the ground truth images $I, D$ for optimization.

The optimization of Gaussian parameters is performed via gradient descent using photometric and geometric supervision from observed keyframes. During optimization, the Gaussians are adjusted to minimize a combination of image reconstruction loss and sparsity-based regularization. This procedure progressively improves the fidelity and coverage of the map while preserving real-time performance.

% \subsection{Gaussian Splatting SLAM}

\subsection{Mapping}

We follow prior work \cite{yugay2023gaussian, liso2024loopy} and represent the scene as a collection of 3DGS submaps where each submap $\mathbf{P}^s_{a,l}$ is defined for agent $a$ as:
\begin{equation}
\mathbf{P}^s_{a,l} = \left\{ G_i^s(\mu, \Sigma, o, c) \mid i = 1, \dots, N \right\},
\end{equation}
where $G_i^s(\mu, \Sigma, o, c)$ represents a point with individual Gaussian mean $\mu \in \mathbb{R}^3$, covariance matrix $\Sigma \in \mathbb{R}^{3\times3}$, opacity value $o \in \mathbb{R}$, and RGB color $c \in \mathbb{R}^3$. 
% Unlike prior work, we define each submap with respect to a local frame $l$ where the first keyframe begins at the identity matrix $I$. 

% Submaps are transformed to the global frame $g$ for reconstruction \todo{fix this}: 

% \begin{equation}
% \mathbf{P}^s_{a,g} = \left\{ T_l^{g} G_{i,l}^s(\mu, \Sigma, o, c) \mid i = 1, \dots, N \right\},
% \end{equation}

Unlike prior work, we define each submap with respect to a local frame $l$, where the first keyframe begins at the identity matrix, $\mathbf{I}$. After pose graph optimization, submaps are transformed to the global frame $g$ using the optimized transformation $T_{a,l}^g$ for agent $a$ and local submap $l$:
\begin{equation}
\mathbf{P}^s_{a,g} = \left\{ T_{a,l}^g \cdot G_{i,l}^s(\mu, \Sigma, o, c) \mid i = 1, \dots, N \right\},
\end{equation}
where $T_{a,l}^g \in SE(3)$ maps the submap from its local frame to the global coordinate frame. This transformation is initialized from odometry and refined through pose graph optimization using both sequential and loop closure constraints, detailed in Section \ref{sec:pgo}.

\subsubsection{Sub-map Initialization}

The first submap $P^s_l$ starts with the first frame $\mathbf{I}^s_f$ where the pose is defined in the local frame $l$ at the origin and models a sequence of keyframes $f \in \{1, \ldots, K\}$. Rather than processing the entire global submap simultaneously, we follow \cite{yugay2023gaussian, liso2024loopy} and initialize new submaps $P^2, \ldots, P^M$ after the current keyframe pose exceeds a translation threshold $d_{\text{max}}$ or rotation threshold $\theta_{\text{max}}$ with respect to the first frame of the submap. We redefine the pose at the first keyframe of each submap $\mathbf{I}^s_f$ as the origin to ensure globally consistent optimization. At any time, only the current active submap which is bounded to a limited size is processed in order to reduce the computational cost of exploring larger scenes.

\subsubsection{Sub-map Building}

For subsequent keyframes, 3D Gaussians are added to newly observed or sparse parts of the active submap. Once the following keyframe pose $\mathbf{I}_k^s$ has been tracked, defining a pose transformation from $T_k^{k-1}$, we use a dense point cloud from the keyframe RGB-D measurement to seed new Gaussians. We sample points uniformly in sparse regions of the map as determined by the rendered opacity of the active map where rendered $\alpha$ values are below a threshold $\alpha_{\text{min}}$. These Gaussians are then optimized using photometric and geometric supervision with gradient descent.

\subsubsection{Sub-map Optimization}

All Gaussian points in the active submap are optimized every time new Gaussians are added based on a fixed number of iterations. We optimize the submap over the rendering loss of depth and color of each keyframe $f$ in the submap after filtering by a soft alpha mask $M_{\alpha}$, a polynomial of the alpha map that suppresses poorly observed or sparsely reconstructed regions and an error inlier mask  $M_\text{in}$, a binary mask that discards pixels where either the color or depth error exceeds a frame-relative threshold:
\begin{align}
\label{eq:render_loss}
\mathcal{L}_{\text{map}} = \sum M_{\text{in}} M_{\alpha} \cdot \big( 
&\lambda_c \lVert \hat{I}^s_j - I^s_j \rVert_{\mathcal{L}1} \nonumber \\
&+ (1-\lambda_c) \lVert \hat{D}^s_j - D^s_j \rVert_{\mathcal{L}1} 
\big),
\end{align}
where $\lambda_c$ weights the color and depth losses, and $\lVert \cdot \rVert_{\mathcal{L}1}$ denotes the $\mathcal{L}_1$ loss between two images.

\subsection{Tracking}
\label{sec:tracking}

\begin{figure}[t!]
  \centering

  \begin{subfigure}[b]{0.49\linewidth}
    \centering
    \includegraphics[width=0.99\linewidth]{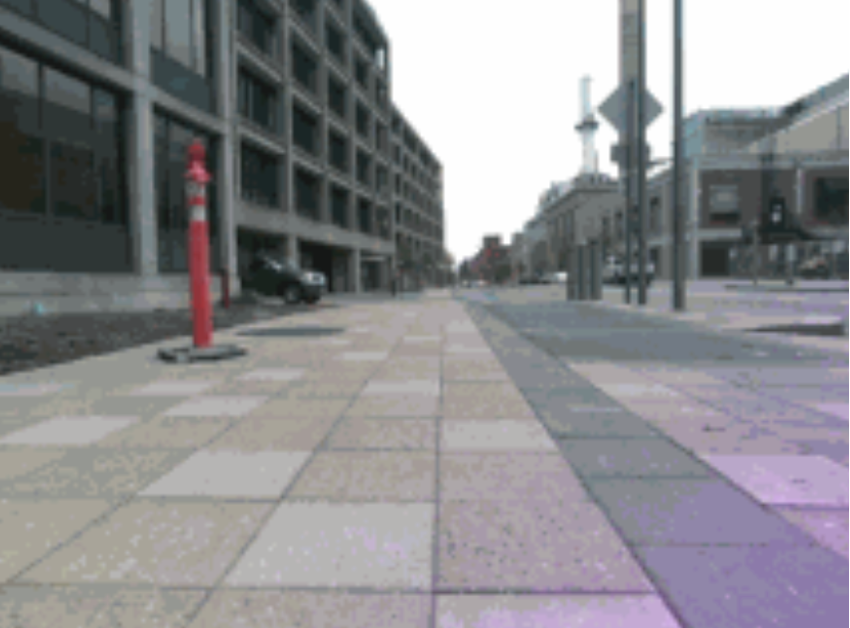}
    % \caption{Before rotation}
    \label{fig:first}
  \end{subfigure}
  % \hfill
  \begin{subfigure}[b]{0.49\linewidth}
    \centering
    \includegraphics[width=0.98\linewidth]{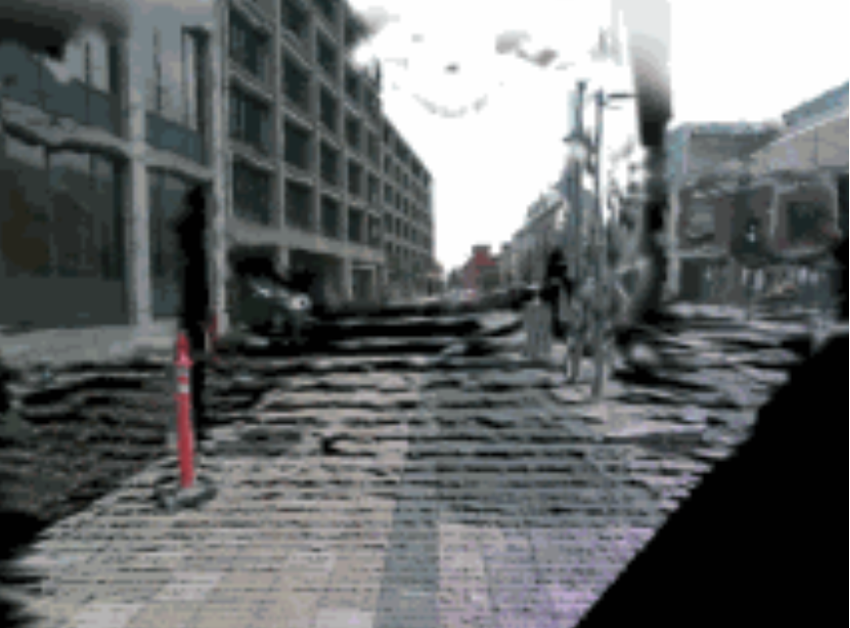}
    % \caption{After rotation}
    \label{fig:2}
  \end{subfigure}

  \caption{Without local optimization, optimizing rotation results in large movements with respect to the origin which may be far from the agent's current position. This example demonstrates the renders of a scene in the Kimera-Multi Outdoor dataset \cite{tian2023resilient} optimization (left) and after rotating with respect to the origin (right).}
  \label{fig:sidebyside}
  \vspace*{-0.1in}
\end{figure}

We perform tracking in two stages: an initial frame-to-frame pose estimation followed by a frame-to-model refinement using rendering-based optimization.
To initialize the current camera pose \( T_i \), we use a coarse relative pose estimate obtained with a relative transformation \( T_{i-1,i} \) found by minimizing a hybrid color-depth odometry objective to provide a fast, coarse initialization for \( T_i \):
\begin{equation}
T_i = T_{i-1} \cdot T_{i-1,i},
\end{equation}
where \( T_{i-1,i} \) is the relative transformation estimated by visual odometry.
In the refinement stage, we separately optimize the rotation and translation components of \( T_i \) in a \textit{local} frame. That is, we express updates relative to the local coordinate system centered at the current pose rather than in the global world frame. This is important because optimizing rotations and translations with respect to the world frame origin that the agent's current pose may be far away from, can result in imbalanced gradients and degraded convergence. We show a qualitative example of degraded rendering performance when optimizing rotation with respect to the world origin in Fig. \ref{fig:sidebyside}.

We formulate the tracking loss \( \mathcal{L}_\text{tracking} \) as the photometric and geometric difference between the input frame and the model rendering under the current pose:
\begin{equation}
\arg\min_{R, \mathbf{t}} \mathcal{L}_\text{track}\left( \hat{I}(R, \mathbf{t}), \hat{D}(R, \mathbf{t}), I, D, \alpha \right),
\end{equation}
where \( R \) and \( \mathbf{t} \) denote the rotation and translation to be optimized, \( \hat{I}(R, \mathbf{t}) \) and \( \hat{D}(R, \mathbf{t}) \) are the rendered color and depth maps produced by transforming the reconstructed model with rotation \( R \) and translation \( \mathbf{t} \), and \( I \) and \( D \) are the input color and depth maps at frame \( i \). The tracking loss follows the rendering loss from Eq. \ref{eq:render_loss} optimized over rotation \( R \) and translation \( \mathbf{t} \). During optimization, all 3D Gaussian parameters of the scene are kept frozen. Only the pose parameters (rotation \( R \) and translation \( \mathbf{t} \)) are updated to minimize the tracking loss.

% The final tracking loss is defined as:
% \begin{equation}
% \mathcal{L}_\text{track} = \sum M_\text{in} \cdot M_\text{$\alpha$} \cdot \left( \lambda_c \left\| \hat{I} - I \right\|_{\mathcal{L}1} + (1 - \lambda_c) \left\| \hat{D} - D \right\|_{\mathcal{L}1} \right),
% \end{equation}
% where \( \lambda_c \) weights the relative importance between color and depth consistency.

% During optimization, all 3D Gaussian parameters of the scene are kept frozen. Only the pose parameters (rotation \( R \) and translation \( \mathbf{t} \)) are updated to minimize the tracking loss.

This two-stage tracking strategy of coarse frame-to-frame hybrid color-depth odometry optimization followed by local frame-to-model optimization enables both fast initialization and accurate alignment even in large, complex environments.

\subsection{Loop Closure Detection}

To ensure global consistency across agent submaps, we incorporate both intra-agent and inter-agent loop closure detection based on keyframe similarity and geometric alignment. Each keyframe stores a NetVLAD descriptor \cite{arandjelovic2016netvlad} computed at capture time associated with its respective submap. These descriptors are stored in a local submap database, and loop closure candidates are detected by querying the current keyframe against this database. For intra-agent loop closures, each agent queries its own submap database, while inter-agent loop closures query databases of other agents when available.

Given a query keyframe descriptor $\mathbf{v}_q$, we retrieve the top-$k$ nearest neighbors $\{\mathbf{v}_j\}_{j=1}^{k}$ from the relevant database using cosine similarity
% \begin{equation}
% \text{sim}(\mathbf{v}_q, \mathbf{v}_j) = \frac{\mathbf{v}_q \cdot \mathbf{v}_j}{\|\mathbf{v}_q\| \|\mathbf{v}_j\|},
% \end{equation}
and accept a match if the similarity exceeds a threshold $\tau_{\text{sim}}$ and the spatial distance between the associated keyframe poses exceeds a minimum baseline  $\Delta{\text{min}}$.

For candidate matches, we estimate an initial transformation between submaps to initialize refinement. In the case of intra-agent loop closures, we use the previously tracked camera poses to compute an initial transform:
\begin{equation}
T_{\text{init}} = T_j^{-1} T_q,
\end{equation}
where $T_q$ and $T_j$ are the estimated poses of the query and match keyframes in the global frame.

For inter-agent loop closures or long-range intra-agent matches where drift may have accumulated, we compute $T_{\text{init}}$ using a coarse geometric alignment method. Specifically, we estimate a relative transformation between the query and match frames using dense RGB-D registration with a hybrid depth-color loss:
\begin{equation}
T_{\text{init}} = \arg\min_T \mathcal{L}_{\text{hybrid}}(T),
\end{equation}
where $\mathcal{L}_{\text{hybrid}}$ jointly minimizes photometric and geometric error between consecutive RGB-D frames using a multi-scale registration pyramid and an iterative solver.

Given an initial transformation $T_{\text{init}}$, we refine the loop closure alignment via point-to-plane ICP on dense colored point clouds of each submap:
\begin{equation}
\mathcal{L}_{\text{ICP}} = \sum_i \left( \left( \mathbf{n}_i^\top ( \mathbf{p}_i^T - \mathbf{q}_i ) \right)^2 \right),
\end{equation}
where $\mathbf{p}_i$ is the source point, $\mathbf{q}_i$ is the target point, and $\mathbf{n}_i$ is the surface normal at the target point. Optimization proceeds until convergence using the following a distance threshold $d_{\text{max}}$, maximum number of iterations $N_{\text{iter}}$ and relative fitness and RMSE thresholds. This produces the final transformation $T^*_{\text{loop}}$, as well as alignment metrics including fitness score $f$ and inlier RMSE $\rho$.

Candidate loop closures are filtered using alignment quality metrics. We accept a loop closure if $f > \tau_f \quad \text{and} \quad \rho < \tau_\rho$ where $\tau_f$ and $\tau_\rho$ are thresholds on fitness and inlier RMSE, respectively. Accepted loop closures are incorporated into a pose graph as additional constraints between submap origins. Let $T_{a,l}^g$ and $T_{b,l}^g$ be the local-to-global transformations of submaps $P^s_{a,l}$ and $P^s_{b,l}$, then the loop closure constraint is added as:
\begin{equation}
T_{a,l}^g T_{ab} = T_{b,l}^g,
\end{equation}
where $T_{ab} = T^*_{\text{loop}}$ is the estimated transformation from submap $b$ to $a$. All constraints are subsequently optimized jointly via pose graph optimization.

\subsection{Pose Graph Optimization}
\label{sec:pgo}

% To collaboratively refine the agents' trajectory estimates, we build a global pose graph to combine all agents' poses and tracking and loop closure measurements in one optimization.

% Without loss of generality, we assign the first pose of the first agent to the origin. Each agent's individual trajectory in the pose graph is initialized using the refined pose from the tracking stage \todo{reference section?}. Intra-agent loop closures are added to the individual trajectories using the full-rank relative pose transformations that come from the loop closure detection process. The agents' trajectories are combined into one global pose graph by using inter-agent loop closures to transform the trajectories into the first agent's trajectory frame. Once the pose graph is formed with all agents' measurements, we perform a batch optimization over the full graph to minimize the total pose error across all agents. In this work, we implement our multi-agent pose graph optimization in GTSAM \todo{cite GTSAM} and use Levenberg-Marquardt optimization, though we note that any nonlinear optimization framework may be used here.

To jointly refine the trajectories of all agents, we construct a global pose graph where nodes correspond to the submap keyframe poses $\{T_{a,f}^g\}$ of agent $a$ in the global frame $g$, and edges represent relative pose constraints from both tracking and loop closure detections.

We assign the first keyframe of the first agent as the fixed origin of the global coordinate frame, denoted as
$T_{0,0}^g = \mathbf{I}_{4 \times 4}$.
Each agent’s trajectory is initialized using the output of the local tracking process described in Section~\ref{sec:tracking}. Intra-agent loop closures provide relative pose constraints between keyframes within the same agent:
\begin{equation}
\mathcal{C}_{\text{intra}} = \left\{ (i, j, \hat{T}_{i,j}, \Sigma_{i,j}) \right\},
\end{equation}
where $\hat{T}_{i,j}$ is the relative transformation estimated by registration and $\Sigma_{i,j}$ is the corresponding information matrix.

Inter-agent loop closures transform one agent’s trajectory into the frame of another, yielding cross-agent constraints:
\begin{equation}
\mathcal{C}_{\text{inter}} = \left\{ (i, j, \hat{T}_{i,j}^{a,b}, \Sigma_{i,j}^{a,b}) \right\},
\end{equation}
with $a \ne b$ and $T_{i,j}^{a,b}$ representing the relative transform from keyframe $i$ of agent $a$ to keyframe $j$ of agent $b$.

The full graph $\mathcal{G}$ is then defined over the union of agent poses and loop closure constraints:
\begin{equation}
\mathcal{G} = (\mathcal{V}, \mathcal{E}) = \left( \left\{ T_i \right\}, \mathcal{C}_{\text{tracking}} \cup \mathcal{C}_{\text{intra}} \cup \mathcal{C}_{\text{inter}} \right).
\end{equation}
We solve for globally consistent poses $\{T_i^*\}$ by minimizing the total error across all relative pose constraints using nonlinear least squares:
\begin{equation}
\mathcal{L}_{\text{graph}} = \sum_{(i,j) \in \mathcal{E}} \left\| \log\left( \hat{T}_{i,j}^{-1} T_i^{-1} T_j \right) \right\|^2_{\Sigma_{i,j}},
\end{equation}
where $\log(\cdot)$ denotes the $\mathfrak{se}(3)$ logarithm map, and the norm is weighted by the information matrix $\Sigma_{i,j}$. We implement the multi-agent pose graph optimization in GTSAM \cite{dellaert2012factor} and use Levenberg-Marquardt optimization, though we note that any nonlinear optimization framework may be used here.

% Once the globally consistent poses $\{T_i^*\}$ are obtained, each local keyframe pose $T_{j}^{(l)}$ and all 3D Gaussian parameters in the corresponding submap are updated. The transformed keyframe pose becomes:
% \begin{equation}
% T_j^{(l)} \leftarrow T_i^* T_j^{(l)},
% \end{equation}
% where $T_j^{(l)}$ is the keyframe's original local pose within submap $i$.
Once the globally consistent submap poses ${T_{a,l}^g}$ are obtained from pose graph optimization, each keyframe pose $T_{j}^{(l)}$ and all associated 3D Gaussian parameters within submap $\mathbf{P}^s_{a,l}$ are transformed into the global frame. The updated global keyframe pose becomes:
\begin{equation}
T_j^{(g)} \leftarrow T_{a,l}^g \cdot T_j^{(l)},
\end{equation}
where $T_j^{(l)}$ is the keyframe's original pose in the local submap frame $l$, and $T_j^{(g)}$ is its pose in the global frame. Similarly, each Gaussian $G_{i,l}^s(\mu, \Sigma, o, c)$ in the submap is transformed as:
\begin{equation}
\mu^{(g)} = R \mu^{(l)} + t, \quad \Sigma^{(g)} = R \Sigma^{(l)} R^\top,
\end{equation}
where $T_{a,l}^g = [R \mid t]$ is the corresponding transformation applied to the Gaussian means and covariances. Color values remain fixed, as we do not recompute spherical harmonics during this update. This produces globally aligned submaps $\mathbf{P}^s_{a,g}$ suitable for joint rendering and evaluation. 

% {\color{red} do you need an algorithm pseudocode?}

% To maintain consistency between the new pose graph and the rendered map, we also transform each Gaussian mean $\mu_j$ and covariance matrix $\Sigma_j$ in the submap using the corrected pose. Specifically,
% \begin{align}
% \mu_j &\leftarrow T_i^* \mu_j, \\
% \Sigma_j &\leftarrow R_i^* \Sigma_j R_i^{*T},
% \end{align}
% where $R_i^*$ is the rotational component of $T_i^*$. Color values remain fixed, as we do not recompute spherical harmonics during this update.

% This update ensures that the geometry of each submap aligns with the globally optimized trajectory, enabling accurate and photorealistic rendering across all agents' reconstructions.

%% file: paper/results.tex
\section{Evaluation}

\subsection{Experimental Setup}

\subsubsection{Datasets}

To evaluate our method against existing multi-agent dense SLAM approaches, we conduct experiments on two complementary datasets that reflect both idealized and real-world conditions.
We first benchmark on the Multiagent Replica dataset \cite{hu2023cp, straub2019replica}, which includes four synthetic indoor sequences featuring two collaborating RGB-D agents. Each sequence offers clean depth and color data without noise, blur, or reflective artifacts. This setting allows for controlled comparison of trajectory accuracy and reconstruction fidelity under ideal conditions.

\input{tables/replica_ate_rmse}
\input{tables/replica_ma_ate_rmse}

To test scalability and robustness in real-world scenarios, we also evaluate on subsets of the Kimera-Multi Outdoor dataset \cite{tian2023resilient}, a large-scale outdoor RGB-D dataset involving six agents with varying trajectory overlap. We evaluate on a combined 1.85 km of sequences individually ranging from 239.25 m to 409.74 m that present significant challenges such as depth sensor noise, reflections, and motion blur. In our evaluation, Agents 1, 2, and 3 refer to the ACL\_Jackal, Hathor and Thoth traverses for the Outside 1 team and the Sparkal1, Sparkal2 and ACL\_Jackal2 traverses for the Outside 2 team respectively. Together, these two datasets allow us to assess performance across both clean and noisy multi-agent environments.

% To benchmark against existing multi-agent dense neural SLAM approaches, we test our method on the Multiagent Replica dataset \todo{cite here} which consists of four two-agent RGB-D traverses in a simulated environment. These sequences are 5m to 9m long \todo{figure out how long these are} and take place in a synthetic indoor environments, where we evaluate the trajectory and rendering performance on a small-scale idealized environment with no depth artifacts or blurred images. We also test our method on subsets of the Kimera-Multi outdoor dataset which contains 6 agents with varying levels of overlap throughout the map. The sequences we evaluate on vary from \todo{length to length} and exhibit challenges related to real-world large-scale datasets including depth artifacts, reflections and motion blur.

\subsubsection{Implementation Details}

We evaluate GRAND-SLAM on a machine with two NVIDIA GeForce RTX 3090 GPUs, each with 24 GB VRAM. During evaluation of multi-agent settings, we run GRAND-SLAM on each agent individually then apply the server module with inter-agent loop closures and multi-agent pose graph optimization separately to evaluate both single-agent and multi-agent performance. 

\subsubsection{Baseline Methods}

We evaluate our system’s tracking and mapping performance against both classical and learning-based SLAM baselines. For single-agent tracking, GRAND-SLAM is compared against representative RGB-D and stereo visual SLAM methods, including ORB-SLAM3 \cite{campos2021orb} and Point-SLAM \cite{sandstrom2023point}, as well as recent Gaussian splatting-based approaches: MonoGS \cite{matsuki2024gaussian}, Gaussian-SLAM \cite{yugay2023gaussian}, and MAGiC-SLAM \cite{yugay2025magic}.

In the collaborative setting, we benchmark tracking performance against multi-agent SLAM systems such as SWARM-SLAM \cite{lajoie2023swarm}, CCM-SLAM \cite{schmuck2019ccm}, CP-SLAM \cite{hu2023cp} and MAGiC-SLAM \cite{yugay2025magic}. 

To assess the quality of dense map reconstruction, we compare rendering fidelity against MAGiC-SLAM \cite{yugay2025magic}, Gaussian-SLAM \cite{yugay2023gaussian} and CP-SLAM \cite{hu2023cp}, which use dense scene representations for single- and multi-agent mapping.

\input{tables/kimera_outdoor_single_agent_tracking}

Since GRAND-SLAM supports both single- and multi-agent operation, with or without loop closures, we report results under each setting to allow for fair comparisons across the full range of baseline systems.

% To evaluate tracking performance, we compare against RGB-D and stereo visual SLAM approaches including ORB-SLAM3 and Point-SLAM, as well as recent gaussian splatting SLAM approaches - MonoGS Gaussian SLAM, and single-agent MAGiC SLAM. For the collaborative setting, we compare against multi-agent SLAM systems including SWARM-SLAM, CCM-SLAM and CP-SLAM. We also compare against a concurrent work that also performs multi-agent gaussian splatting SLAM, Magic-SLAM, which is tailored for indoor environments. We evaluate the rendering performance against MAGiC-SLAM and CP-SLAM to compare against dense neural scene representations used in multi-agent SLAM. 

% GRAND-SLAM can operate with or without loop closures in a single agent setting and in a multi agent setting, so we include evaluations against single-agent SLAM approaches as well.

\subsection{Camera Tracking}

Table \ref{tab:replica_sa_rmse} compares GRAND-SLAM to state of the art single-agent visual SLAM approaches. In these cases, we report the average of the two agents' performances for each dataset. In the case of MAGiC-SLAM, Table \ref{tab:replica_sa_rmse} includes the single agent results. GRAND-SLAM achieves superior tracking accuracy than other methods without loop closure and with intra-agent loop closure, we are consistent with or outperform baseline approaches in all cases. 

We also compare against multi-agent approaches on the Multiagent Replica dataset in Table \ref{table:mar}. GRAND-SLAM is consistent with or marginally outperforms MAGiC-SLAM, while outperforming all other baseline methods. This demonstrates that GRAND-SLAM is capable of running at high performance in small-scale indoor settings while also generalizing to large-scale challenging real-world datasets. 

On the Kimera-Multi dataset, we evaluate against visual SLAM approaches to demonstrate the current limitations of the state of the art large-scale outdoor settings using real-world datasets, shown in Table \ref{table:km_track}. We note that both MAGiC-SLAM and Gaussian SLAM failed partway through the Agent 2 traverse of the Outside 2 Team, so partial results are included. MAGiC-SLAM also fails to consistently find correct loop closures in these settings, so single-agent results are reported. GRAND-SLAM outperforms all baseline methods on the Kimera-Multi dataset with and without loop closures.

\begin{figure*}[ht]
    \centering
    \includegraphics[width=\linewidth, trim={1cm, 4.5cm, 1.5cm, 0}, clip]{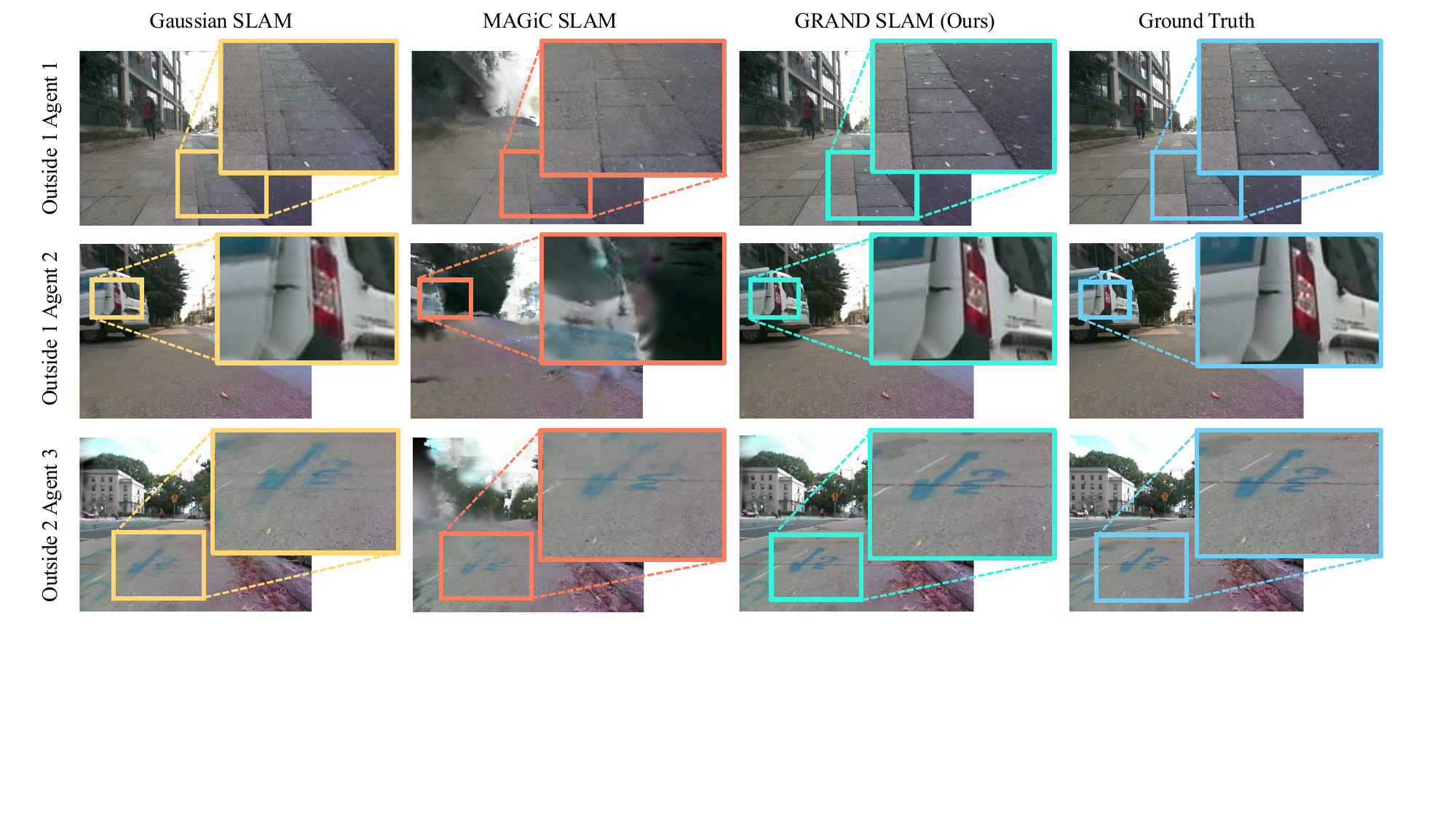}
    \caption{Renders from the Kimera-Multi dataset demonstrate GRAND-SLAM's superior performance compared to baseline methods in large-scale scenes where it maintains photorealistic rendering results.}
    \label{fig:renders}
    \vspace*{-0.1in}
\end{figure*}

\input{tables/kimera_outdoor_render}

\input{tables/replica_outdoor_rendering}

\subsection{Rendering}

We evaluate the rendering performance on the Multiagent Replica dataset of submaps compared to MAGiC-SLAM and CP-SLAM, two multiagent neural SLAM approaches, in Table \ref{table:mar-render}. GRAND-SLAM significantly outperforms baseline approaches on training view synthesis in the small-scale indoor setting. We also evaluate the rendering performance on the Kimera-Multi outdoor dataset of Gaussian SLAM and MAGiC SLAM in Table \ref{table:km-render}, where GRAND-SLAM outperforms all baseline methods as well. We note that MAGiC-SLAM and Gaussian-SLAM both failed partway through one traverse, so partial results prior to the failure are reported. GRAND-SLAM has superior rendering performance than baseline methods in both small-scale indoor and large-scale outdoor settings.

%% file: tables/replica_ate_rmse.tex
\begin{table}[t!]
\centering
\caption{Single-Agent tracking performance on Multiagent Replica (ATE RMSE ↓ [cm]), where colors denote \colorbox{myteal}{\textbf{first}} \colorbox{mycoral}{second} and \colorbox{myyellow}{third} best performance.  }
\begin{tabular}{@{}lcccccc@{}}
\toprule
\textbf{Method}& \textbf{Off0} & \textbf{Apt0} & \textbf{Apt1} & \textbf{Apt2} & \textbf{Avg}  \\ \midrule
ORB-SLAM3 \cite{campos2021orb} & 0.60 & 1.07 & 4.94 & 1.36 & 1.99  \\
Gaussian-SLAM \cite{yugay2023gaussian} & \cellcolor[HTML]{F8CA8C}0.33  & 0.41 & 30.13 & 121.96 & 38.21 \\
MonoGS \cite{matsuki2024gaussian} & \cellcolor[HTML]{FAF3B3}0.38  & \cellcolor[HTML]{95CEC7}\textbf{0.21}  & 3.33         & 0.54          & 1.15                 \\
MAGiC-SLAM \cite{yugay2025magic}      & 0.42         & 0.38        & \cellcolor[HTML]{FAF3B3}0.54         & \cellcolor[HTML]{FAF3B3}0.66          & \cellcolor[HTML]{FAF3B3}0.50               \\
\textbf{GRAND-SLAM (w/o LC)}         & 0.44         & \cellcolor[HTML]{FAF3B3}0.28         & \cellcolor[HTML]{F8CA8C}0.46         & \cellcolor[HTML]{F8CA8C}0.27          & \cellcolor[HTML]{F8CA8C}0.36           \\
\textbf{GRAND-SLAM}&  \cellcolor[HTML]{95CEC7}\textbf{0.32} & \cellcolor[HTML]{F8CA8C}0.23 & \cellcolor[HTML]{95CEC7}\textbf{0.35} & \cellcolor[HTML]{95CEC7}\textbf{0.17} & \cellcolor[HTML]{95CEC7}\textbf{0.27}  \\ 
\bottomrule
\label{tab:replica_sa_rmse}
\end{tabular}\vspace*{-0.1in}
\end{table}

%% file: tables/replica_ma_ate_rmse.tex
\begin{table}[b!]
\centering
\caption{Multi-Agent Tracking performance on Multiagent Replica (ATE RMSE ↓ [cm]), where colors denote \colorbox{myteal}{\textbf{first}} \colorbox{mycoral}{second} and \colorbox{myyellow}{third} best performance. }
\begin{tabular}{@{}llccccc@{}}
\toprule
\textbf{Method} & \textbf{Agent} & \textbf{O-0} & \textbf{A-0} & \textbf{A-1} & \textbf{A-2} & \textbf{Avg} \\ \midrule
CCM-SLAM \cite{schmuck2019ccm} & \textbf{Agt 1} & 9.84 & \redx & 2.12 & 0.51 & -\\
Swarm-SLAM \cite{lajoie2023swarm} &  & 1.07 & 1.61 & 4.62 & 2.69 & 2.50  \\
CP-SLAM \cite{hu2023cp} &  & 0.50 & 0.62 & 1.11 & 1.41 & 0.91 \\
MAGiC-SLAM (w/o LC) &  & 0.44 & 0.30 & 0.48 & 0.91 & 0.53 \\
MAGiC-SLAM \cite{yugay2025magic} &  & 0.31 & 0.13 & \textbf{0.21} & 0.42 & 0.27 \\ 
\textbf{GRAND-SLAM (w/o LC)} &  & \textbf{0.27} & \textbf{0.27} & 0.47 & 0.33 & 0.34 \\
\textbf{GRAND-SLAM} &  & 0.28 & \textbf{0.27} & 0.28 & \textbf{0.18} & \textbf{0.25} \\\midrule
CCM-SLAM \cite{schmuck2019ccm} & \textbf{Agt 2} & 0.76 & \redx & 9.31 & 0.48 & -\\
Swarm-SLAM \cite{lajoie2023swarm} &  & 1.76 & 1.98 & 6.50 & 8.53 & 4.69 \\
CP-SLAM \cite{hu2023cp} &  & 0.79 & 1.28 & 1.72 & 2.41 & 1.55 \\
MAGiC-SLAM (w/o LC) &  & 0.41 & 0.46 & 0.61 & 0.41 & 0.47 \\
MAGiC-SLAM \cite{yugay2025magic} &  & \textbf{0.24} & 0.21 & \textbf{0.30} & 0.22 & \textbf{0.24} \\ 
\textbf{GRAND-SLAM (w/o LC)} &  & 0.43 & 0.22 & 0.44 & 0.20 & 0.32 \\
\textbf{GRAND-SLAM} &  & 0.25 & \textbf{0.19} & 0.36 & \textbf{0.18} & 0.25 \\\midrule
CCM-SLAM \cite{schmuck2019ccm} & \textbf{Avg} & 5.30 & \redx & 5.71 & 0.49 & -\\
Swarm-SLAM \cite{lajoie2023swarm} &  & 1.42 & 1.80 & 5.56 & 5.61 & 3.60  \\
CP-SLAM \cite{hu2023cp} &  & 0.65 & 0.95 & 1.42 & 1.91 & 1.23 \\
MAGiC-SLAM (w/o LC) &  & 0.42 & 0.38 & 0.54 & 0.66 & 0.50 \\
MAGiC-SLAM \cite{yugay2025magic} &  & \cellcolor[HTML]{F8CA8C}0.28 & \cellcolor[HTML]{95CEC7}\textbf{0.17} & \cellcolor[HTML]{95CEC7}\textbf{0.26} & \cellcolor[HTML]{FAF3B3}0.32 & \cellcolor[HTML]{F8CA8C}0.26 \\ 
\textbf{GRAND-SLAM (w/o LC)} &  & \cellcolor[HTML]{FAF3B3}0.44 & \cellcolor[HTML]{FAF3B3}0.28 & \cellcolor[HTML]{FAF3B3}0.46 & \cellcolor[HTML]{F8CA8C}0.27  & \cellcolor[HTML]{FAF3B3}0.36 \\
\textbf{GRAND-SLAM} &  & \cellcolor[HTML]{95CEC7}\textbf{0.27} & \cellcolor[HTML]{F8CA8C}0.23 & \cellcolor[HTML]{F8CA8C}0.32 & \cellcolor[HTML]{95CEC7}\textbf{0.18} & \cellcolor[HTML]{95CEC7}\textbf{0.25} \\
\bottomrule
\label{table:mar}
\end{tabular}
\end{table}

%% file: tables/kimera_outdoor_single_agent_tracking.tex
\begin{table}[t!]
\centering
\caption{Tracking Performance on Kimera-Multi Outdoor (ATE RMSE ↓ [m]), where colors denote \colorbox{myteal}{\textbf{first}} and \colorbox{mycoral}{second} best performance, and partial results from failed runs are denoted with an asterisk* and averages including partial runs are shown in \textcolor{gray}{gray}.}
\begin{tabular}{@{}llccccc@{}}
\toprule
\textbf{Method} & \textbf{Agent} & \textbf{Outside 1} & \textbf{Outside 2} & \textbf{Avg} \\ \midrule
ORB-SLAM3 \cite{campos2021orb} & \textbf{Agt 1} & 14.11 & \textbf{2.72} & 8.42 \\
Gaussian SLAM \cite{johari2023eslam} &  & 356.61 & 71.16 & 213.89 \\
MAGiC-SLAM \cite{yugay2025magic}  &  & 24.13 & 11.13 & 17.63  \\
\textbf{GRAND-SLAM (w/o LC)} &  & 6.43 & 10.19 & 8.31 \\
\textbf{GRAND-SLAM} &  & \textbf{3.95} & 10.93 & \textbf{7.44}
\\
\midrule
ORB-SLAM3 \cite{campos2021orb} & \textbf{Agt 2} & \textbf{7.07} & 13.12 & 10.10 \\
Gaussian SLAM \cite{johari2023eslam} &  & 119.68 & 7.66* & \textcolor{gray}{63.67} \\
% fails at 1831
MAGiC-SLAM \cite{yugay2025magic}  &  & 98.33 & 10.50* &  \textcolor{gray}{54.42} \\
\textbf{GRAND-SLAM (w/o LC)} &  & 9.74 & 4.63 & 7.19 \\
\textbf{GRAND-SLAM} &  & 8.93 & \textbf{4.54} & \textbf{6.74} \\
\midrule
ORB-SLAM3 \cite{campos2021orb} & \textbf{Agt 3} & 7.48 & 18.99 & 13.24 \\
Gaussian SLAM \cite{johari2023eslam} & & 1150.42  & 195.95 & 673.19 \\
MAGiC-SLAM \cite{yugay2025magic}  &  & 172.81 & 47.86 & 110.34  \\
\textbf{GRAND-SLAM (w/o LC)} &  & \textbf{1.05} &7.93 & 4.49 \\
\textbf{GRAND-SLAM} &  & 1.30 & \textbf{6.25} & \textbf{3.78} \\
% ORB-SLAM3 \cite{campos2021orb} & \textbf{Agt 1} & 14.11 & 18.99 & x \\

% Gaussian SLAM \cite{johari2023eslam} &  & 356.61 & 195.95 & 276.28 \\
% MAGiC SLAM \cite{yugay2024magic}  &  & 24.13 & 47.86 & 36.00   \\
% Ours (w/o LC) &  & 6.43 & 7.93 & 7.18 \\
% Ours &  & \textbf{3.95} & \textbf{6.25} & \textbf{5.10}
% \\
% \midrule
% ORB-SLAM3 \cite{campos2021orb} & \textbf{Agt 2} & \textbf{7.07} & 13.12 & x \\
% Gaussian SLAM \cite{johari2023eslam} &  & 119.68 & \redx & x \\
% MAGiC SLAM \cite{yugay2024magic}  &  & 98.33 & \redx & x  \\
% Ours (w/o LC) &  & 9.74 & 4.63 & - \\
% Ours &  & 8.93 & \textbf{4.54} & - \\
% \midrule
% ORB-SLAM3 \cite{campos2021orb} & \textbf{Agt 3} & 7.48 & \textbf{2.72} & x \\
% Gaussian SLAM \cite{johari2023eslam} & & 1150.42  & 71.16 & -\\
% MAGiC SLAM \cite{yugay2024magic}  &  & 172.81 & 11.13 & -  \\
% Ours (w/o LC) &  & \textbf{1.05} & 10.19 & - \\
% Ours &  & 1.30 & 10.93 & - \\
% ORB-SLAM3 \cite{campos2021orb} & \textbf{Agt 2} & 7.07 & 2.72 & x \\
% Gaussian SLAM \cite{johari2023eslam} &  & 119.68 & 71.16 & 95.42 \\
% MAGiC SLAM \cite{yugay2024magic}  &  & 98.33 & 11.13 & 54.73  \\
% Ours (w/o LC) &  & 9.74 & 10.19 & 9.97 \\
% Ours &  & 8.93 & 10.93& 9.93 \\
% \midrule
% ORB-SLAM3 \cite{campos2021orb} & \textbf{Agt 3} & 7.48 & 13.12 & x \\
% Gaussian SLAM \cite{johari2023eslam} & & 1150.42  & \redx & -\\
% MAGiC SLAM \cite{yugay2024magic}  &  & 172.81 & \redx & -  \\
% Ours (w/o LC) &  & 1.05 & 4.63 & 2.84 \\
% Ours &  & 1.30 & 4.54 & 2.92 \\
\midrule
ORB-SLAM3 \cite{campos2021orb} & \textbf{Avg} & 9.55 & 11.61 & 10.58 \\
Gaussian SLAM \cite{johari2023eslam} &  & 542.24 & \textcolor{gray}{91.59} & \textcolor{gray}{316.92}\\
MAGiC-SLAM \cite{yugay2025magic}  &  & 98.42 & \textcolor{gray}{23.16} & \textcolor{gray}{60.79}   \\
\textbf{GRAND-SLAM (w/o LC)} &  & \cellcolor[HTML]{F8CA8C}5.74 & \cellcolor[HTML]{F8CA8C}7.58 & \cellcolor[HTML]{F8CA8C}6.66 \\
\textbf{GRAND-SLAM} &  & \cellcolor[HTML]{95CEC7}\textbf{4.73} & \cellcolor[HTML]{95CEC7}\textbf{7.24 } & \cellcolor[HTML]{95CEC7}\textbf{4.99} \\
% CCM-SLAM \cite{yugay2024magic} & \textbf{Agt 2} & 0.76 & \redx \\
% Swarm-SLAM \cite{yugay2024magic} &  & 1.76 & 1.98 & 6.50 & 8.53 & 4.69 \\
% CP-SLAM \cite{yugay2024magic} &  & 0.79 & 1.28 & 1.72 & 2.41 & 1.55 \\
% MAGiC-SLAM (w/o LC) &  & 0.41 & 0.46 & 0.61 & 0.41 & 0.47 \\
% MAGiC-SLAM &  & \textbf{0.24} & 0.21 & \textbf{0.30} & 0.22 & \textbf{0.24} \\ 
% CORGI-SLAM (w/o LC) &  & 0.43 & 0.22 & 0.44 & 0.20 & 0.32 \\
% CORGI-SLAM &  & 0.25 & \textbf{0.19} & 0.36 & \textbf{0.18} & 0.25 \\\midrule
% CCM-SLAM \cite{yugay2024magic} & \textbf{Avg} & 5.30 & \redx & 5.71 & 0.49 & -\\
% Swarm-SLAM \cite{yugay2024magic} &  & 1.42 & 1.80 & 5.56 & 5.61 & 3.60  \\
% CP-SLAM \cite{yugay2024magic} &  & 0.65 & 0.95 & 1.42 & 1.91 & 1.23 \\
% MAGiC-SLAM (w/o LC) &  & 0.42 & 0.38 & 0.54 & 0.66 & 0.50 \\
% MAGiC-SLAM &  & \cellcolor[HTML]{D6F5E4}0.28 & \cellcolor[HTML]{5FCF97}\textbf{0.17} & \cellcolor[HTML]{5FCF97}\textbf{0.26} & \cellcolor[HTML]{FAF3B3}0.32 & \cellcolor[HTML]{D6F5E4}0.26 \\ 
% CORGI-SLAM (w/o LC) &  & \cellcolor[HTML]{FAF3B3}0.44 & \cellcolor[HTML]{FAF3B3}0.28 & \cellcolor[HTML]{FAF3B3}0.46 & \cellcolor[HTML]{D6F5E4}0.27  & \cellcolor[HTML]{FAF3B3}0.36 \\
% CORGI-SLAM &  & \cellcolor[HTML]{5FCF97}\textbf{0.27} & \cellcolor[HTML]{D6F5E4}0.23 & \cellcolor[HTML]{D6F5E4}0.32 & \cellcolor[HTML]{5FCF97}\textbf{0.18} & \cellcolor[HTML]{5FCF97}\textbf{0.25} \\
\bottomrule
\label{table:km_track}
\end{tabular}
\end{table}

%% file: tables/kimera_outdoor_render.tex
\begin{table}[t!]
\centering
\caption{Training view synthesis performance on Kimera-Multi dataset, where best performance is shown in \textbf{bold}, partial results from failed runs are denoted with an asterisk* and averages including partial runs are shown in \textcolor{gray}{gray}.}
\begin{tabular}{p{1.5cm} p{1.5cm} c c c}
\toprule
\textbf{Methods} & \textbf{Metrics} & \textbf{Outside 1} & \textbf{Outside 2} & \textbf{Avg} \\
\midrule
% \multirow{4}{*}{Gaussian-SLAM \cite{yugay2023gaussian}} 
& PSNR  ↑ & 24.59& 24.31*& \textcolor{gray}{24.45}  \\
Gaussian & SSIM ↑      & 0.90  & 0.89* & \textcolor{gray}{0.90} \\
SLAM & LPIPS ↓     & 0.18  & 0.17*  & \textcolor{gray}{0.18}  \\
& Depth L1 ↓ & 1.19  & 1.45* & \textcolor{gray}{1.32}  \\
\midrule
% \multirow{4}{*}{MAGiC-SLAM \cite{yugay2025magic}} 
& PSNR  ↑ & 16.12 & 15.63* & \textcolor{gray}{15.88}  \\
MAGiC & SSIM ↑      & 0.49  & 0.50* & \textcolor{gray}{0.50} \\
SLAM & LPIPS ↓     & 0.54  & 0.53*  &  \textcolor{gray}{0.54}  \\
& Depth L1 ↓ & 3.86  & 4.71* & \textcolor{gray}{4.29}  \\
\midrule
% \multirow{4}{*}{\textbf{GRAND-SLAM}} 
& PSNR  ↑ & 28.48 & 26.62 & \textbf{27.44}  \\
\textbf{GRAND} & SSIM ↑      & 0.97  & 0.96 & \textbf{0.97} \\
\textbf{SLAM} & LPIPS ↓     & 0.10  & 0.12  & \textbf{0.11}  \\
& Depth L1 ↓ & 1.17  & 1.61 & \textbf{1.39}  \\
\bottomrule
\end{tabular}
\label{table:km-render}
\vspace*{-0.1in}
\end{table}

%% file: tables/replica_outdoor_rendering.tex
\begin{table}[t!]
\centering
\caption{Training view synthesis performance on Multiagent Replica dataset, where best performance is shown in \textbf{bold}. }
\begin{tabular}{l l c c c c c}
\toprule
\textbf{Methods} & \textbf{Metrics} & \textbf{O-0} & \textbf{A-0} & \textbf{A-1} & \textbf{A-2} & \textbf{Avg} \\
\midrule
% \multirow{4}{*}{CP-SLAM} 
& PSNR  ↑ & 28.56 & 26.12 & 12.16 & 23.98 & 22.71 \\
CP & SSIM ↑      & 0.87  & 0.79  & 0.31  & 0.81  & 0.69 \\
SLAM & LPIPS ↓     & 0.29  & 0.41  & 0.97  & 0.39  & 0.51 \\
& Depth L1 ↓ & 2.74  & 19.93 & 66.77 & 2.47  & 22.98 \\
\midrule
% \multirow{4}{*}{MAGiC-SLAM} 
& PSNR  ↑ & 39.32 & 36.96 & 30.01 & 30.73 & 34.26 \\
MAGiC & SSIM ↑      & 0.99  & 0.98  & 0.95  & 0.96  & 0.97 \\
SLAM & LPIPS ↓     & 0.05  & 0.09  & 0.18  & 0.17  & 0.12 \\
& Depth L1 ↓ & 0.41  & 0.64  & 3.16  & 0.99  & 1.30 \\
\midrule
% \multirow{4}{*}{\textbf{GRAND-SLAM}} 
& PSNR  ↑ & 43.12 & 44.15 & 38.65 & 39.46& \textbf{41.35} \\
\textbf{GRAND} & SSIM ↑  & 0.99  & 0.99 & 0.99 & 0.99  & \textbf{0.99} \\
\textbf{SLAM} & LPIPS ↓ &  0.03 & 0.03& 0.05  & 0.05 &  \textbf{0.04} \\
& Depth L1 ↓ & 0.25  & 0.31 & 0.77 & 0.29  & \textbf{0.41} \\
\bottomrule
\end{tabular}
\vspace*{-0.1in}
\label{table:mar-render}
\end{table}

% Apart0 p1: psnr': 43.9320200874485, 'lpips': 0.024754056936142656, 'ssim': 0.9955998941543162, 'depth_l1_train_view': 0.002374337878977723
% apar0 p2: {'psnr': 44.37499726493404, 'lpips': 0.026224360262401075, 'ssim': 0.9957151015599569, 'depth_l1_train_view': 0.003740737236405584, 'num_renders': 477}

% apar1 part 1: {'psnr': 38.020089616007326, 'lpips': 0.05470801300729656, 'ssim': 0.9924146284757288, 'depth_l1_train_view': 0.00990015938091284, 'num_renders': 509}
% apart1 part2: {'psnr': 37.59040570354082, 'lpips': 0.047421354570414916, 'ssim': 0.9903673053500187, 'depth_l1_train_view': 0.00542388588423837, 'num_renders': 502}

% apart2 part1: {'psnr': 39.28673600200638, 'lpips': 0.053518332451938155, 'ssim': 0.9929602469106119, 'depth_l1_train_view': 0.002831772545610454, 'num_renders': 502}
% part 2: {'psnr': 39.6339519898256, 'lpips': 0.046634028346570794, 'ssim': 0.9938714649968731, 'depth_l1_train_view': 0.0030568773017912534, 'num_renders': 499}

% office part1: {'psnr': 43.465389105552, 'lpips': 0.027040930200810633, 'ssim': 0.9956208869415225, 'depth_l1_train_view': 0.0024870418286900182, 'num_renders': 261}
% {'psnr': 42.783097432454426, 'lpips': 0.031955450319995485, 'ssim': 0.9957509922981262, 'depth_l1_train_view': 0.0025511129794176667, 'num_renders': 300}

%% file: paper/conclusion.tex
\section{Conclusion}

We present GRAND-SLAM, a multi-agent dense SLAM approach using 3DGS as the underlying scene representation, designed for large-scale outdoor environments. In future work, we plan to integrate compression techniques to better manage communication and memory constraints for larger scenes. GRAND-SLAM outperforms state of the art neural and 3DGS SLAM methods in rendering and tracking performance, showing promise as a method using local optimization to scale effectively for large-scale photorealistic SLAM. 

%% file: root.bbl
\begin{thebibliography}{10}
\providecommand{\url}[1]{#1}
\csname url@samestyle\endcsname
\providecommand{\newblock}{\relax}
\providecommand{\bibinfo}[2]{#2}
\providecommand{\BIBentrySTDinterwordspacing}{\spaceskip=0pt\relax}
\providecommand{\BIBentryALTinterwordstretchfactor}{4}
\providecommand{\BIBentryALTinterwordspacing}{\spaceskip=\fontdimen2\font plus
\BIBentryALTinterwordstretchfactor\fontdimen3\font minus \fontdimen4\font\relax}
\providecommand{\BIBforeignlanguage}[2]{{%
\expandafter\ifx\csname l@#1\endcsname\relax
\typeout{** WARNING: IEEEtran.bst: No hyphenation pattern has been}%
\typeout{** loaded for the language `#1'. Using the pattern for}%
\typeout{** the default language instead.}%
\else
\language=\csname l@#1\endcsname
\fi
#2}}
\providecommand{\BIBdecl}{\relax}
\BIBdecl

\bibitem{cadena2016past}
C.~Cadena, L.~Carlone, H.~Carrillo, Y.~Latif, D.~Scaramuzza, J.~Neira, I.~Reid, and J.~J. Leonard, ``Past, present, and future of simultaneous localization and mapping: Toward the robust-perception age,'' \emph{IEEE Transactions on robotics}, vol.~32, no.~6, pp. 1309--1332, 2016.

\bibitem{fuentes2015visual}
J.~Fuentes-Pacheco, J.~Ruiz-Ascencio, and J.~M. Rend{\'o}n-Mancha, ``Visual simultaneous localization and mapping: a survey,'' \emph{Artificial intelligence review}, vol.~43, pp. 55--81, 2015.

\bibitem{kazerouni2022survey}
I.~A. Kazerouni, L.~Fitzgerald, G.~Dooly, and D.~Toal, ``A survey of state-of-the-art on visual slam,'' \emph{Expert Systems with Applications}, vol. 205, p. 117734, 2022.

\bibitem{davison2007monoslam}
A.~J. Davison, I.~D. Reid, N.~D. Molton, and O.~Stasse, ``Monoslam: Real-time single camera slam,'' \emph{IEEE transactions on pattern analysis and machine intelligence}, vol.~29, no.~6, pp. 1052--1067, 2007.

\bibitem{klein2007parallel}
G.~Klein and D.~Murray, ``Parallel tracking and mapping for small ar workspaces,'' in \emph{2007 6th IEEE and ACM international symposium on mixed and augmented reality}.\hskip 1em plus 0.5em minus 0.4em\relax IEEE, 2007, pp. 225--234.

\bibitem{mur2017orb}
R.~Mur-Artal and J.~D. Tard{\'o}s, ``{Orb-SLAM2}: An open-source {SLAM} system for monocular, stereo, and {RGB-D} cameras,'' \emph{IEEE transactions on robotics}, vol.~33, no.~5, pp. 1255--1262, 2017.

\bibitem{schops2019bad}
T.~Schops, T.~Sattler, and M.~Pollefeys, ``Bad {SLAM}: Bundle adjusted direct {RGB-D} {SLAM},'' in \emph{Proceedings of the IEEE/CVF CVPR}, 2019, pp. 134--144.

\bibitem{whelan2015elasticfusion}
T.~Whelan, S.~Leutenegger, R.~F. Salas-Moreno, B.~Glocker, and A.~J. Davison, ``Elasticfusion: Dense {SLAM} without a pose graph.'' in \emph{Robotics: science and systems}, vol.~11.\hskip 1em plus 0.5em minus 0.4em\relax Rome, Italy, 2015, p.~3.

\bibitem{newcombe2011kinectfusion}
R.~A. Newcombe, S.~Izadi, O.~Hilliges, D.~Molyneaux, D.~Kim, A.~J. Davison, P.~Kohi, J.~Shotton, S.~Hodges, and A.~Fitzgibbon, ``Kinectfusion: Real-time dense surface mapping and tracking,'' in \emph{2011 10th IEEE international symposium on mixed and augmented reality}.\hskip 1em plus 0.5em minus 0.4em\relax Ieee, 2011, pp. 127--136.

\bibitem{sucar2021imap}
E.~Sucar, S.~Liu, J.~Ortiz, and A.~J. Davison, ``Imap: Implicit mapping and positioning in real-time,'' in \emph{Proceedings of the IEEE/CVF international conference on computer vision}, 2021, pp. 6229--6238.

\bibitem{zhu2022nice}
Z.~Zhu, S.~Peng, V.~Larsson, W.~Xu, H.~Bao, Z.~Cui, M.~R. Oswald, and M.~Pollefeys, ``Nice-{SLAM}: Neural implicit scalable encoding for {SLAM},'' in \emph{Proceedings of the IEEE/CVF CVPR}, 2022, pp. 12\,786--12\,796.

\bibitem{johari2023eslam}
M.~M. Johari, C.~Carta, and F.~Fleuret, ``E{SLAM}: Efficient dense {SLAM} system based on hybrid representation of signed distance fields,'' in \emph{Proceedings of the IEEE/CVF CVPR}, 2023, pp. 17\,408--17\,419.

\bibitem{wang2023co}
H.~Wang, J.~Wang, and L.~Agapito, ``Co-{SLAM}: Joint coordinate and sparse parametric encodings for neural real-time {SLAM},'' in \emph{Proceedings of the IEEE/CVF CVPR}, 2023, pp. 13\,293--13\,302.

\bibitem{mildenhall2021nerf}
B.~Mildenhall and P.~P. e.~a. Srinivasan, ``{NERF:} representing scenes as neural radiance fields for view synthesis,'' \emph{Communications of the ACM}, vol.~65, no.~1, pp. 99--106, 2021.

\bibitem{kerbl20233d}
B.~Kerbl, G.~Kopanas, T.~Leimk{\"u}hler, and G.~Drettakis, ``3d {Gaussian} splatting for real-time radiance field rendering.'' \emph{ACM Trans. Graph.}, vol.~42, no.~4, pp. 139--1, 2023.

\bibitem{matsuki2024gaussian}
H.~Matsuki, R.~Murai, P.~H. Kelly, and A.~J. Davison, ``{Gaussian} splatting {SLAM},'' in \emph{Proceedings of the IEEE/CVF CVPR}, 2024, pp. 18\,039--18\,048.

\bibitem{yugay2023gaussian}
V.~Yugay, Y.~Li, T.~Gevers, and M.~R. Oswald, ``{Gaussian}-{SLAM}: Photo-realistic dense {SLAM} with {G}aussian splatting,'' \emph{arXiv preprint arXiv:2312.10070}, 2023.

\bibitem{keetha2024splatam}
N.~Keetha, J.~Karhade, K.~M. Jatavallabhula, G.~Yang, S.~Scherer, D.~Ramanan, and J.~Luiten, ``{SplaTAM:} splat track \& map {3D} {Gaussian}s for dense {RGB-D} slam,'' in \emph{IEEE CVPR}, 2024, pp. 21\,357--21\,366.

\bibitem{liso2024loopy}
L.~Liso, E.~Sandstr{\"o}m, V.~Yugay, L.~Van~Gool, and M.~R. Oswald, ``Loopy-{SLAM}: Dense neural {SLAM} with loop closures,'' in \emph{Proceedings of the IEEE/CVF Conference on Computer Vision and Pattern Recognition}, 2024, pp. 20\,363--20\,373.

\bibitem{zhu2024loopsplat}
L.~Zhu, Y.~Li, E.~Sandstr{\"o}m, S.~Huang, K.~Schindler, and I.~Armeni, ``Loopsplat: Loop closure by registering {3D} {G}aussian splats,'' \emph{arXiv preprint arXiv:2408.10154}, 2024.

\bibitem{xu2024glc}
Z.~Xu, Q.~Li, C.~Chen, X.~Liu, and J.~Niu, ``{Glc-SLAM}: {Gaussian} splatting {SLAM} with efficient loop closure,'' \emph{arXiv preprint arXiv:2409.10982}, 2024.

\bibitem{wu2024hgs}
K.~Wu, K.~Zhang, Z.~Zhang, M.~Tie, S.~Yuan, J.~Zhao, Z.~Gan, and W.~Ding, ``{HGS}-mapping: Online dense mapping using hybrid {Gaussian} representation in urban scenes,'' \emph{IEEE RA-L}, 2024.

\bibitem{hong2024liv}
S.~Hong, J.~He, X.~Zheng, C.~Zheng, and S.~Shen, ``{LIV-GaussMap: LiDAR-inertial-visual fusion for real-time {3D} radiance field map rendering},'' \emph{IEEE Robotics and Automation Letters}, 2024.

\bibitem{campos2021orb}
C.~Campos, R.~Elvira, J.~J.~G. Rodr{\'\i}guez, J.~M. Montiel, and J.~D. Tard{\'o}s, ``Orb-{SLAM}3: An accurate open-source library for visual, visual--inertial, and multimap {SLAM},'' \emph{IEEE transactions on robotics}, vol.~37, no.~6, pp. 1874--1890, 2021.

\bibitem{fu2021fast}
Q.~Fu, H.~Yu, X.~Wang, Z.~Yang, Y.~He, H.~Zhang, and A.~Mian, ``Fast orb-slam without keypoint descriptors,'' \emph{IEEE transactions on image processing}, vol.~31, pp. 1433--1446, 2021.

\bibitem{gomez2019pl}
R.~Gomez-Ojeda, F.-A. Moreno, D.~Zuniga-No{\"e}l, D.~Scaramuzza, and J.~Gonzalez-Jimenez, ``{PL-SLAM:} a stereo {SLAM} system through the combination of points and line segments,'' \emph{IEEE Transactions on Robotics}, vol.~35, no.~3, pp. 734--746, 2019.

\bibitem{li2021rgb}
Y.~Li, R.~Yunus, N.~Brasch, N.~Navab, and F.~Tombari, ``{RGB-D} {SLAM} with structural regularities,'' in \emph{2021 IEEE international conference on Robotics and automation (ICRA)}.\hskip 1em plus 0.5em minus 0.4em\relax IEEE, 2021, pp. 11\,581--11\,587.

\bibitem{gong2021planefusion}
B.~Gong and Z.~e.~a. Zhu, ``Planefusion: Real-time indoor scene reconstruction with planar prior,'' \emph{IEEE Transactions on Visualization and Computer Graphics}, vol.~28, no.~12, pp. 4671--4684, 2021.

\bibitem{niessner2013real}
M.~Nie{\ss}ner, M.~Zollh{\"o}fer, S.~Izadi, and M.~Stamminger, ``Real-time {3D} reconstruction at scale using voxel hashing,'' \emph{ACM Transactions on Graphics (ToG)}, vol.~32, no.~6, pp. 1--11, 2013.

\bibitem{curless1996volumetric}
B.~Curless and M.~Levoy, ``A volumetric method for building complex models from range images,'' in \emph{Proceedings of the 23rd annual conference on Computer graphics and interactive techniques}, 1996, pp. 303--312.

\bibitem{dai2017bundlefusion}
A.~Dai, M.~Nie{\ss}ner, M.~Zollh{\"o}fer, S.~Izadi, and C.~Theobalt, ``Bundlefusion: Real-time globally consistent {3D} reconstruction using on-the-fly surface reintegration,'' \emph{ACM Transactions on Graphics (ToG)}, vol.~36, no.~4, p.~1, 2017.

\bibitem{newcombe2011dtam}
R.~A. Newcombe, S.~J. Lovegrove, and A.~J. Davison, ``Dtam: Dense tracking and mapping in real-time,'' in \emph{2011 ICCV}.\hskip 1em plus 0.5em minus 0.4em\relax IEEE, 2011, pp. 2320--2327.

\bibitem{stuhmer2010real}
J.~St{\"u}hmer, S.~Gumhold, and D.~Cremers, ``Real-time dense geometry from a handheld camera,'' in \emph{Joint Pattern Recognition Symposium}.\hskip 1em plus 0.5em minus 0.4em\relax Springer, 2010, pp. 11--20.

\bibitem{yang2022vox}
X.~Yang, H.~Li, H.~Zhai, Y.~Ming, Y.~Liu, and G.~Zhang, ``Vox-fusion: Dense tracking and mapping with voxel-based neural implicit representation,'' in \emph{2022 IEEE International Symposium on Mixed and Augmented Reality (ISMAR)}.\hskip 1em plus 0.5em minus 0.4em\relax IEEE, 2022, pp. 499--507.

\bibitem{sandstrom2023point}
E.~Sandstr{\"o}m, Y.~Li, L.~Van~Gool, and M.~R. Oswald, ``Point-{SLAM}: Dense neural point cloud-based {SLAM},'' in \emph{IEEE ICCV}, 2023, pp. 18\,433--18\,444.

\bibitem{ha2024rgbd}
S.~Ha, J.~Yeon, and H.~Yu, ``{RGBD GS-ICP {SLAM}},'' \emph{arXiv preprint arXiv:2403.12550}, 2024.

\bibitem{zhu2024robust}
Z.~Zhu, Y.~Fang, X.~Li, C.~Yan, F.~Xu, C.~Yuen, and Y.~Li, ``Robust {Gaussian} splatting {SLAM} by leveraging loop closure,'' \emph{arXiv preprint arXiv:2409.20111}, 2024.

\bibitem{yugay2025magic}
V.~Yugay, T.~Gevers, and M.~R. Oswald, ``Magic-slam: Multi-agent gaussian globally consistent slam,'' in \emph{Proceedings of the Computer Vision and Pattern Recognition Conference}, 2025, pp. 6741--6750.

\bibitem{schmuck2019ccm}
P.~Schmuck and M.~Chli, ``{CCM-SLAM: Robust and efficient centralized collaborative monocular simultaneous localization and mapping for robotic teams},'' \emph{Journal of Field Robotics}, vol.~36, no.~4, pp. 763--781, 2019.

\bibitem{karrer2018cvi}
M.~Karrer, P.~Schmuck, and M.~Chli, ``{CVI-SLAM—collaborative visual-inertial SLAM},'' \emph{IEEE Robotics and Automation Letters}, vol.~3, no.~4, pp. 2762--2769, 2018.

\bibitem{lajoie2023swarm}
P.-Y. Lajoie and G.~Beltrame, ``Swarm-{SLAM}: Sparse decentralized collaborative simultaneous localization and mapping framework for multi-robot systems,'' \emph{IEEE Robotics and Automation Letters}, vol.~9, no.~1, pp. 475--482, 2023.

\bibitem{hu2023cp}
J.~Hu, M.~Mao, H.~Bao, G.~Zhang, and Z.~Cui, ``{CP-SLAM}: Collaborative neural point-based {SLAM} system,'' \emph{Advances in Neural Information Processing Systems}, vol.~36, pp. 39\,429--39\,442, 2023.

\bibitem{tian2023resilient}
Y.~Tian, Y.~Chang, L.~Quang, A.~Schang, C.~Nieto-Granda, J.~P. How, and L.~Carlone, ``Resilient and distributed multi-robot visual {SLAM}: Datasets, experiments, and lessons learned,'' in \emph{2023 IEEE/RSJ International Conference on Intelligent Robots and Systems (IROS)}.\hskip 1em plus 0.5em minus 0.4em\relax IEEE, 2023, pp. 11\,027--11\,034.

\bibitem{arandjelovic2016netvlad}
R.~Arandjelovic, P.~Gronat, A.~Torii, T.~Pajdla, and J.~Sivic, ``Netvlad: Cnn architecture for weakly supervised place recognition,'' in \emph{IEEE CVPR}, 2016, pp. 5297--5307.

\bibitem{dellaert2012factor}
F.~Dellaert, ``Factor graphs and {GTSAM:} a hands-on introduction,'' \emph{Georgia Institute of Technology, Tech. Rep}, vol.~2, no.~4, 2012.

\bibitem{straub2019replica}
J.~Straub, T.~Whelan, L.~Ma, Y.~Chen, E.~Wijmans, S.~Green, J.~J. Engel, R.~Mur-Artal, C.~Ren, S.~Verma \emph{et~al.}, ``The replica dataset: A digital replica of indoor spaces,'' \emph{arXiv preprint arXiv:1906.05797}, 2019.

\end{thebibliography}
